\documentclass[conference]{IEEEtran}
\IEEEoverridecommandlockouts
\usepackage[utf8]{inputenc}
\usepackage[T5]{fontenc}

\usepackage{cite}
\usepackage{hyperref}
\usepackage{amsmath,amssymb,amsfonts}
\usepackage{algorithmic}
\usepackage{graphicx}
\usepackage{textcomp}
\usepackage{xcolor}
\usepackage{authblk}
\usepackage{longtable}
\usepackage{float}
\usepackage{makecell}
\usepackage{pgfplots}
\usepackage{pgf-pie}
\usepackage{tikz}
\usepackage{amsmath}
\usepackage[font={small},justification=centering]{caption} 
\usepackage{threeparttable}
\usepackage{multirow}
\usepackage{footnote}
\usepackage{hyperref}
\usepackage{flushend}

\usetikzlibrary{patterns}

\def\BibTeX{{\rm B\kern-.05em{\sc i\kern-.025em b}\kern-.08em
    T\kern-.1667em\lower.7ex\hbox{E}\kern-.125emX}}

\newcommand{\footref}[1]{%
    $^{\ref{#1}}$%
}

\makeatletter
\@ifpackageloaded{hyperref}{%
  \renewcommand\footref[1]{%
    \begingroup 
    \unrestored@protected@xdef\@thefnmark{%
      \ref*{#1}%
    }%
    \endgroup 
    \ifHy@hyperfootnotes 
       \expandafter\@firstoftwo 
    \else 
       \expandafter\@secondoftwo 
    \fi 
    {\hyperref[#1]{\strut\H@@footnotemark}}{\@footnotemark}%
  }%
}{}%

\newcommand\savedlabel{}%
\AtBeginDocument{\let\savedlabel=\label}%
\newcommand\footnotereflabel[1]{%
   \@bsphack
   \begingroup
   \def\@currentHref{Hfootnote.\theHfootnote}\savedlabel{#1}%
   \endgroup
   \@esphack
}%

\makeatother

\begin{document}


\title{\textcolor{black}{An Experimental Study of Deep Neural Network Models for Vietnamese Multiple-Choice Reading Comprehension} \\}

\author[1,2,*]{Son T. Luu}
\author[1,2,*]{Kiet Van Nguyen}
\author[1,2,*]{Anh Gia-Tuan Nguyen}
\author[1,2,*]{Ngan Luu-Thuy Nguyen}
\affil[1]{University of Information Technology, Ho Chi Minh City, Vietnam}
\affil[2]{Vietnam National University, Ho Chi Minh City, Vietnam}

\affil[ ]{Email: *\textit {\{sonlt,kietnv,anhngt,ngannlt\}@uit.edu.vn}}

\maketitle

\begin{abstract}
\textcolor{black}{Machine reading comprehension (MRC) is a challenging task in natural language processing that makes computers understanding natural language texts and answer questions based on those texts. There are many techniques for solving this problems, and word representation is a very important technique that impact most to the accuracy of machine reading comprehension problem in the popular languages like English and Chinese. However, few studies on MRC have been conducted in low-resource languages such as Vietnamese. In this paper, we conduct several experiments on neural network-based model to understand the impact of word representation to the Vietnamese multiple-choice machine reading comprehension. Our experiments include using the Co-match model on six different Vietnamese word embeddings and the BERT model for multiple-
choice reading comprehension. On the ViMMRC corpus, the accuracy of BERT model is 61.28\% on test set. }

\end{abstract}


\section{Introduction}
\label{section1}
Machine reading comprehension (MRC) is a very interesting and challenging problem in Natural Language Processing (NLP). It aims to make the computer can read and understand the texts and answer the question from the texts. Multiple-choice reading comprehension is a sub-problem of machine reading comprehension, in which machines answer the question by predicting the correct answer from a list of answer options. In machine reading comprehension problems, the corpus plays a very important role, which can impact the precision of the result. Some of famous corpus such as MCTest{\cite{richardson-etal-2013-mctest}}, CNN/Daily Mail\cite{hermann2015teaching} and RACE{\cite{lai2017large}} was created for multiple-choice reading comprehension problem.

For Vietnamese language, there are two corpora such as ViMMRC \cite{nguyen2020vimmrc} and ViNewsQA \cite{van2020new}. However, we focus on multiple-choice machine reading comprehension in this paper. Thus, the ViMMRC corpus\footnote{The corpus can be downloaded at \url{https://sites.google.com/uit.edu.vn/uit-nlp/} for research purposes.} by Nguyen et at. {\cite{nguyen2020vimmrc}} was created for multiple choice reading comprehension. This corpus contains a total of 417 reading texts, each text has at least 5 multiple choices question. In this paper, we will conduct experment on the ViMMRC dataset with Co-matching model - a deep neural model for multiple-choice reading comprehension \cite{wang-etal-2018-co} on six different word embeddings provided by Vu et al.\cite{vu:2019n} and the BERT model for multiple-choice reading comprehension \cite{devlin-etal-2019-bert, Si2019WhatDB}. Then, we compare the results to find the model and the word embedding which received the highest results on the ViMMRC corpus. Finally, we also make the error analysis on each word embedding to explore the impact of word embeddings on the results. 

Our main contributions in this paper are:
\begin{itemize}
\item \textcolor{black}{First, we implement the Co-match model and BERT model on ViMMRC with six different Vietnamese word embeddings. Then, we compare the accuracy of these word embedding to find out which is the best one for the ViMMRC corpus.}

\item \textcolor{black}{Second, we makes the error analysis between these word embeddings on Co-match model and between Co-match model with BERT model to study their effect on the result of multiple choices reading comprehension task.}

\end{itemize}

The paper content is structured as follows. Section \ref{section2} reviews related corpora and models for machine reading comprehension. Section \ref{section3} takes an overview of the ViMMRC corpus. Section \ref{section4} presents our approaches for Vietnamese machine reading comprehension. Section \ref{section5} describes our experiments and error analysis results on the ViMMRC corpus. Finally, Section \ref{section6} concludes the paper.
\section{Related work}
\label{section2}
First, we briefly introduced several famous corpora used for the study of multiple-choice reading comprehension, described as follows.
\begin{itemize}
\item MCTest created by Richardson et al. is an open-domain machine reading comprehension corpus. It contains 500 stories and 2000 questions which are carefully limited for young children who can read and understand the texts \cite{richardson-etal-2013-mctest}.
\item CNN/Daily Mail created by Hermann et al. is a corpus which combines 2 different corpora: CNN and Daily Mail, in which 93,000 articles for CNN and 220,000 articles for Daily Mail and has nearly 1.4 million questions \cite{hermann2015teaching}.
\item RACE created by Lai et al. is another large scale reading comprehension corpus for examination. The corpus contains nearly 28,000 passages and 100,000 questions generated by English instructors, which covers a variety of topics for testing the ability of reading and understanding of students. \cite{lai2017large}.
\item The ViMMRC corpus \cite{nguyen2020vimmrc} created by Nguyen et al. contains 417 paragraphs and more than 2,000 questions from Vietnamese textbook from 1${st}$ to 5${th}$ grade.
\end{itemize}
Second, we introduce two popular models used for machine-reading comprehension:
\begin{itemize}
\item Co-match introduced by Wang et al. is a deep neural model that used for multiple-choice reading comprehension. This model received the best performance on the RACE dataset, which is 50.4\% \cite{wang-etal-2018-co}.
\item BERT invented by Devlin et al. \cite{devlin-etal-2019-bert} is a state-of-the-arts model now for language understanding since BERT is conceptually simple, but empirically powerful.

\end{itemize}
Finally, on the Vietnamese language, Vu et al. \cite{vu:2019n} propose a process that used for extracting, evaluating and visualizing multiple pre-trained word embeddings. Besides, the authors also provided six pre-trained word embeddings that trained from Wikipedia in Vietnamese.

\section{The Corpus}
\label{section3}
\subsection{Corpus statistic}
\begin{table}[H]
\caption{EXAMPLES OF READING COMPREHENSION IN THE CORPUS \ \cite{nguyen2020vimmrc}}
\label{table1}
\begin{center}
\begin{tabular}{|p{8.3cm}|}
\hline
\\
\textbf{Passage:} \\
Ngay giữa sân trường, sừng sững một cây bàng. Mùa đông, cây vươn dài những cành khẳng khiu, trụi lá. Xuân sang, cành trên cành dưới chi chít những lộc non mơn mởn. Hè về, những tán lá xanh um che mát một khoảng sân trường. Thu đến, từng chùm quả chín vàng trong kẽ lá. \\
(\textit{\textbf{English:} In the middle of the school- yard stood a towering tropical almond tree. In winter, the tree stretches out its slender, leafless branches. As spring arrives, its branches on the branches below are spangled with young buds. Summer approaches and its green foliage shades the yard. Autumn comes, re- vealing bunches of gold ripen fruits dangling in its leaves}) \\
\\
\textbf{Questions:} \\
1) Cây bàng được trồng ở đâu?
(\textit{\textbf{English:} Where is the tropical almond tree planted?})\\
A. Ngay giữa sân trường (\textit{\textbf{English:} In the middle of the schoolyard}). \\
B. Trồng ở ngoài đường (\textit{\textbf{English:} In the street}). \\
C. Gần sông (\textit{\textbf{English:} Near the river}). \\
D. Dưới mái hiên trường (\textit{\textbf{English:} Under the porch}). \\
\\
2) Những bộ phận nào của cây được nhắc đến trong bài đọc? (\textit{\textbf{English:} What parts of the tree are mentioned?}) \\
A. Cành và lá (\textit{\textbf{English:} Branches and leaves}). \\
B. Lá và quả (\textit{\textbf{English:} Leaves and fruit}). \\
C. Cành, lá, lộc, tán lá và quả (\textit{\textbf{English:} Branches, leaves, buds, foliage and fruit}). \\
D. Lộc, quả và tán cây (\textit{\textbf{English:} Buds, fruit and foliage}). \\
\\
\textbf{Answers:} \\
1.A 2.C \\
\\
\hline
\end{tabular}
\end{center}
\end{table}

As mentioned in Section \ref{section1}, the ViMMRC corpus contains 417 texts from the Vietnamese Student Textbook from Grade 1 to Grade 5. In the corpus, each passage has at least 5 corresponding questions and each question has 4 choices. Table \ref{table1} shows an example of the reading text in the corpus.

Besides, we also categorized the text to each grade. The statistical information about the corpus on each grade includes the number of texts, the number of questions, and vocabulary size. Vocabulary size is calculated by the number of words in texts, questions, and options. In this paper, we use the Pyvi tool\footnote{Pyvi tool: \url{https://pypi.org/project/pyvi/}} for word segmentation. Table \ref{table2} shows statistical information about the ViMMRC corpus.

\begin{table}[H]
\caption{STATISTIC OF THE VIMMRC CORPUS BY EACH GRADE}
\label{table2}
\begin{center}
\begin{tabular}{|c|c|c|c|}
\hline
\textbf{Grade} & \textbf{Num. texts} & \textbf{Num. questions} & \textbf{Vocab. size} \\
\hline
1 & 10 & 60 & 595 \\ 
\hline
2 & 70 & 514 & 3,325 \\
\hline
3 & 188 & 759 & 4,666 \\
\hline 
4 & 99 & 709 & 5,006 \\
\hline
5 & 120 & 741 & 5,702 \\
\hline
\textbf{Total} & \textbf{417} & \textbf{2,783} & \textbf{10,099} \\
\hline

\end{tabular}
\end{center}
\end{table}

The difficulty of the texts is based on the amount of reading texts, questions and vocabulary size that children can read and understand. As described from Table \ref{table2}, the difficulty of reading texts is increased by each grade, grade 1 is the easiest level and grade 5 is the most difficult level. 

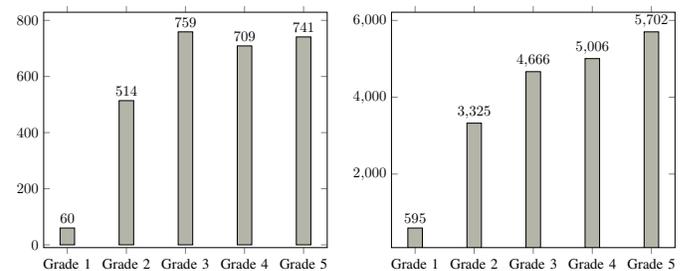
\begin{figure}[H]
\begin{minipage}[t]{.24\textwidth}
    \definecolor{grey}{HTML}{b4b4a9}
    \begin{tikzpicture}[scale=0.55]
        \begin{axis}[
            ybar,
            symbolic x coords={Grade 1, Grade 2, Grade 3, Grade 4, Grade 5},
            xtick=data,
            nodes near coords,
            nodes near coords align={vertical},
    	    ylabel near ticks,
        ]
        \addplot[black, fill=grey] coordinates {
        (Grade 1, 60) (Grade 2, 514) (Grade 3, 759) (Grade 4, 709) (Grade 5, 741)
        };
        \end{axis}
    \end{tikzpicture}
\end{minipage}
\begin{minipage}[t]{.2\textwidth}
    \definecolor{grey}{HTML}{b4b4a9}
    \begin{tikzpicture}[scale=0.55]
        \begin{axis}[
            ybar,
            symbolic x coords={Grade 1, Grade 2, Grade 3, Grade 4, Grade 5},
            xtick=data,
            nodes near coords,
            nodes near coords align={vertical},
    	    ylabel near ticks,
        ]
        \addplot[black, fill=grey] coordinates {
        (Grade 1, 595) (Grade 2, 3325) (Grade 3, 4666) (Grade 4, 5006) (Grade 5, 5702)
        };
        \end{axis}
    \end{tikzpicture}
\end{minipage}
\caption{Number of questions and vocabulary size of the ViMMRC corpus by each grade.}
\label{figure:num_question}
\end{figure}

Fig. \ref{figure:num_question} illustrates the number of questions and vocabulary size respectively by each grade. According to Fig. \ref{figure:num_question}, although grade 3 have more question than grade 4 and grade 5, the vocabulary size of grade 3 is lower than grade 4 and grade 5. 

\subsection{Reasoning types}
The types of reasoning of questions directly influenced the ability of understanding the reading texts and extracting the answers from the reading texts. In the ViMMRC corpus, there are 5 reasoning methods on questions \cite{nguyen2020vimmrc}:
\begin{itemize}
\item \textbf{Word matching (WM):} The words in the answer matched exactly from the words in the reading texts. This is the easiest kind of question for the answer.
\item \textbf{Paraphrasing (P):} The question is paraphrased from a sentence in the reading texts by using synonyms, idioms or other representations to describe the same meaning.
\item \textbf{Single-sentence reasoning (SSR):} The answer is inferred from one sentence in the reading texts.
\item \textbf{Multi-sentence reasoning (MSR):} The answer is inferred from more than two sentences in the reading texts. Sometimes people need to summarize information from multiple sentences to receive the answer.
\item \textbf{Ambiguous or insufficient (AB):} The question has many different answers or has no answer at all. 
\end{itemize}

Table \ref {table3} gives an overview of the proportions of each reasoning types in the ViMMRC corpus:
\begin{table}[H]
\caption{REASONING TYPES IN VIMMRC CORPUS}
\label{table3}
\begin{center}
\begin{tabular}{|c|c|c|}
\hline
\textbf{Reasoning type} & \textbf{No. questions} & \textbf{Proportion (\%)} \\
\hline
\textbf{Word matching} & 170 & 6.11 \\
\hline
\textbf{Paraphrasing} & 184 & 6.61 \\
\hline 
\textbf{Single-sentence reasoning} & 797 & 28.64 \\
\hline
\textbf{Multi-sentence reasoning} & 1341 & 48.19 \\
\hline
\textbf{Ambiguous or insufficient} & 291 & 10.46 \\
\hline
\end{tabular}
\end{center}
\end{table}


According to Table \ref{table3}, multi-sentence reasoning takes a large proportion of reasoning types in the ViMMRC corpus. This means many questions in the corpus need to infer from many sentences from the reading texts to find the correct answer. In contrast, word matching - the easiest reasoning types only take a small part in the corpus. 

\section{Methodologies}
\label{section4}
\textcolor{black}{In this section, we introduce the models used for our experiments. Our chosen models are the Co-match model by Wang et al.\cite{wang-etal-2018-co} training with six different Vietnamese pre-trained word embedding provided by Vu et al. \cite{vu:2019n} and BERT model by Devlin et al. \cite{devlin-etal-2019-bert}}

\subsection{Co-matching model}
The Co-match model by Wang et al. \cite{wang-etal-2018-co} is a joint model in which the reading text (passage) can match with both the question and the answer. The model is defined by the set of three elements: \{P, Q, A\}, in which P represents for passages - the reading texts, Q represents for questions and A represents for answers. Denote that:
\begin{itemize}
\item P = $\{p_1, p_2, ..., p_m\}$, where $p_i \in$ P$ (i = 1..m)$ is a sentence represented by an embedding vector with dimension \textit{d}, \textit{m} is the length of P.
\item Q = $\{q_1, q_2, ..., q_n\}$, where $q_j \in$ Q$ (j = 1..n)$ is a sentence represented by an embedding vector with dimension \textit{d}, \textit{n} is the length of Q.
\item A = $\{a_1, a_2, ..., a_k\}$, where $a_q \in$ A$ (q = 1..k)$ is a sentence represented by an embedding vector with dimension \textit{d}, \textit{k} is the length of A.
\end{itemize}
According to \cite{wang-etal-2018-co}, the Co-match model received the accuracy of 50.4\% for the RACE dataset. 

\subsection{Word embedding}
Word embedding is a feature learning technique in NLP by mapping words or phrases from the set of vocabularies to a vector of real numbers. The distributed representations of words, which called word embedding help to improve the accuracy of various natural language models \cite{hoang2017}. \\
In our experiment, we use six Vietnamese pre-trained embeddings\footnote{Vietnamese pre-trained word embedding: \url{https://github.com/vietnlp/etnlp}} provided by Vu et al. \cite{vu:2019n}. Those pre-trained word embedding models are based on those following model: W2V \cite{mikolov2013}, W2V\_C2V \cite{kim2016}, FastText \cite{bojanowski-etal-2017-enriching}, Bert\_Base \cite{devlin-etal-2019-bert}, ELMO \cite{peters-etal-2018-deep}, and MULTI \cite{vu:2019n}. All were trained on Vietnamese Wikipedia texts. The information about the dimension of those word embedding is described in Table \ref{table4}.

\begin{table}[H]
\caption{VIETNAMESE PRE-TRAINED WORD EMBEDDING INFORMATION}
\label{table4}
\begin{center}
\begin{tabular}{|c|c|c|}
\hline
\textbf{Word embedding model} & \textbf{Size} & \textbf{Dimension} \\
\hline
\textit{W2V} & 16,690 & 300 \\
\hline
\textit{\makecell{W2V\_C2V}} & 16,690 & 300 \\
\hline
\textit{fastText} & 15,974 & 300 \\
\hline 
\textit{\makecell{Bert\_Base}} & 14,905 & 768 \\
\hline
\textit{ELMO} & 21,930 & 1,024 \\
\hline
\textit{MULTI} & 21,591 & 2,392 \\
\hline
\end{tabular}
\end{center}
\end{table}

\subsection{\textcolor{black}{BERT}}
BERT \cite{devlin-etal-2019-bert} is a multi-layer bidirectional transformer encoder. In this multiple choices tasks, we built the BERT model by feeding a list of sequences include P - sequence of passages, Q - sequence of questions and A - sequence of right answer choices through BERT hiding layer state. Denote that:
\begin{itemize}
\item $H_p=BERT(P)$ is a sequence of the passage P that have been encoded by BERT hidden states.
\item $H_q=BERT(Q)$ is a sequence of the question Q that have been encoded by BERT hidden states.
\item $H_a=BERT(A)$ is a sequence of right answer A that have been encoded by BERT layer. The right answer here corresponds with the right choice of a question in the ViMMRC corpus.
\end{itemize}

\textcolor{black}{In our experiment, we use the bert-base-multilingual-uncased and bert-base-multilingual-cased pre-trained models for training on ViMMRC corpus. The bert-base-multilingual-uncased pre-trained model has 12-layers and 110 million parameters, trained with lower-cased text on large Wikipedia corpus with 104 different languages, including Vietnamese. The bert-base-multilingual-cased is the same, but was trained on cased text.}
\section{Experiments }
\label{section5}
\subsection{Data preparation}
Based on the corpus as described in Section \ref{section3}, we use the training, development, and test sets of the ViMMRC dataset \cite{nguyen2020vimmrc}. The information about training, development and test set are described in Table \ref{table5}.

\begin{table}[H]
\caption{TRAINING CORPUS STATISTICS}
\label{table5}
\begin{center}
\begin{tabular}{|c|c|c|c|}
\hline
\textbf{} & \textbf{Train} & \textbf{Dev} & \textbf{Test} \\
\hline
\textbf{Number of texts} & 292 & 42 & 83 \\
\hline
\textbf{Number of questions} & 1,975 & 294 & 514 \\
\hline
\textbf{Vocabulary size} & 8,422 & 2,876 & 4,502 \\
\hline
\end{tabular}
\end{center}
\end{table}

In addition, Table \ref{table6} illustrates the proportion of reasoning types of questions on training, development and test sets. It can be inferred from the table that the single-sentence reasoning and the multi-sentences reasoning take the majority proportion on training, development and test sets of the dataset.  

\begin{table}[H]
\caption{REASONING TYPES OF TRAINING CORPUS}
\label{table6}
\begin{center}
\begin{tabular}{|c|c|c|c|}
\hline
\textbf{Reasoning types} & \textbf{Train (\%)} & \textbf{Dev (\%)} & \textbf{Test (\%)} \\
\hline
Word matching & 6.13 & 25.85 & 4.67 \\
\hline
Paraphrasing & 7.29 & 13.95 & 3.89 \\
\hline 
Single-sentence reasoning & 27.65 & 17.35 & 29.18 \\
\hline
Multi-sentence reasoning & 49.52 & 36.73 & 47.67 \\
\hline
Ambiguous or insufficient & 9.42 & 6.12 & 14.59 \\
\hline
\end{tabular}
\end{center}
\end{table}

In addition, after dividing the corpus into training, development and test sets, we apply the pre-processing techniques as below: 

\begin{itemize}
\item Firstly, we segment the reading texts into sentences with the nltk sent\_tokenize tool\footnote{\url{https://pypi.org/project/nltk/}}.
\item After that, we tokenize each sentence in reading texts into words by the pyvi tool\footnote{\footnotereflabel{pyvi}\url{https://pypi.org/project/pyvi/}}.
\item Then, we tokenize the questions into words using the pyvi tool\footref{pyvi}.
\item Finally, we tokenize the answer texts into words using the pyvi tool\footref{pyvi}.
\end{itemize}

After the pre-processing process, we counts the number of vocabularies that appeared in each word embedding. Table \ref{table7} describes the appearance of words in the ViMMRC corpus on each word embedding.

\begin{table}[H]
\caption{NUMBER OF WORDS IN THE VIMMRC CORPUS APPEARED IN EACH WORD EMBEDDING}
\label{table7}
\begin{center}
\begin{tabular}{|c|c|c|}
\hline
\textbf{Word embedding} & \textbf{Appeared words} \\
\hline
\textit{W2V} &  6,430\\
\hline
\textit{\makecell{W2V\_C2V}} &  6,430\\
\hline
\textit{FastText} &  6,380\\
\hline 
\textit{\makecell{Bert\_Base}} & 6,187 \\
\hline
\textit{ELMO} &  6,220 \\
\hline
\textit{MULTI} &  6,220 \\
\hline
\end{tabular}
\end{center}
\end{table}

\subsection{Empirical configuration}
In our experiments, we trained the Co-match model with 30 epochs, batch size equal to 8, dropout ratio equal to 0.2 and learning rate equal to 0.002.

Besides, we set the maximum sequence length of texts equal to 320, training batch size equal to 8, and test batch size equal to 1 for the BERT model. Then, we execute the BERT model with 3 epochs and the learning rate equal to $10^{-5}$.


\subsection{Empirical results}
Table \ref{table8} shows the empirical results of the Co-match model and the BERT model on the ViMMRC corpus.
\begin{table}[H]
\caption{EMPIRICAL RESULTS OF CO-MATCH MODEL ON DIFFERENT WORD EMBEDDING}
\label{table8}
\begin{center}
\begin{tabular}{|c|m{2cm}|c|c|}
\hline
\textbf{Model} & \textbf{Word embeddings} & \textbf{\makecell{Dev \\ Accuracy (\%)}} & \textbf{\makecell{Test \\ Accuracy (\%)}} \\
\hline
\multirow{6}{*}{Co-match} & W2V & 31.63 & 37.35 \\
\cline{2-4}
& W2V\_C2V & 31.63 & 37.35 \\
\cline{2-4}
& FastText & 42.17 & 40.66 \\
\cline{2-4}
& \textbf{Bert\_Base} & \textbf{43.19} & 40.07\\
\cline{2-4}
& \textbf{ELMO} & 42.17 & \textbf{45.17} \\
\cline{2-4}
& MULTI & 38.77 & 42.80 \\
\hline
\multirow{4}{*}{\textcolor{black}{BERT}} & \textbf{\textcolor{black}{bert-base-multilingual-cased}} & \textbf{\textcolor{black}{68.02}} & 60.50 \\
\cline{2-4}
& \textbf{\textcolor{black}{bert-base-multilingual-uncased}} & 65.98 & \textbf{\textcolor{black}{61.28}} \\
\hline
\end{tabular}
\end{center}
\end{table}

According to Table \ref{table8}, for the Co-match model, ELMO gained the highest accuracy on the test set with 45.17\% and Bert\_Base gained the highest accuracy on the development set with 43.19\%. On development set, FastText and ELMO have the same accuracy, which are 42.17\% for both, however, the result is different in the test set, in which EMLO is higher than FastText. Besides, W2V and W2V\_C2V have the same accuracy, and their accuracy on the test set are higher than on the development set. \textcolor{black}{For the BERT model, the result is significantly better than the Co-match model. The bert-base-multilingual-cased model has the highest result on the development set while the bert-base-multilingual-uncased model obtained the highest result on the test set.}

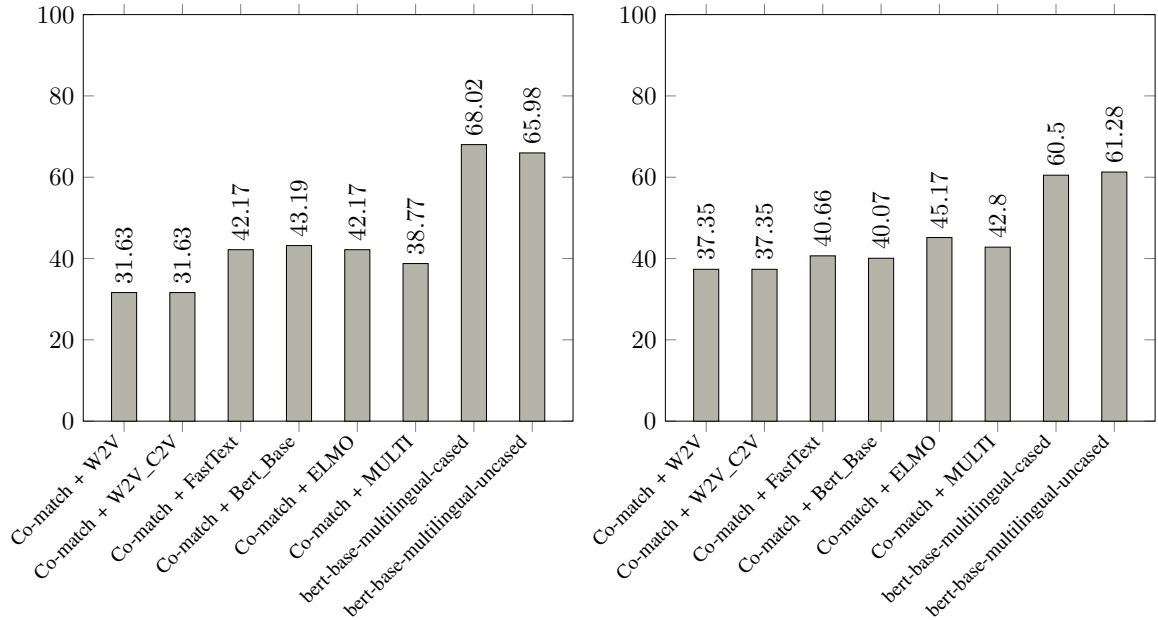
\begin{figure*}[ht]
\centering
\begin{minipage}[t]{.42\textwidth}
    \definecolor{grey}{HTML}{b4b4a9}
    \begin{tikzpicture}[scale=0.95]
        \begin{axis}[
            ybar,
            symbolic x coords={
                Co-match + W2V, 
                Co-match + W2V\_C2V, 
                Co-match + FastText, 
                Co-match + Bert\_Base, 
                Co-match + ELMO, 
                Co-match + MULTI,
                bert-base-multilingual-cased,
                bert-base-multilingual-uncased
            },
            ymin = 0, ymax = 100,
            xtick=data,
            nodes near coords,
            nodes near coords align={vertical},
            every node near coord/.append style={rotate=90, anchor=west},
    	    ylabel near ticks,
    	    x tick label style={font=\footnotesize,rotate=45, anchor=east},
        ]
        \addplot[black, fill=grey] coordinates {
            (Co-match + W2V, 31.63) 
            (Co-match + W2V\_C2V, 31.63) 
            (Co-match + FastText,42.17) 
            (Co-match + Bert\_Base, 43.19) 
            (Co-match + ELMO, 42.17) 
            (Co-match + MULTI, 38.77)
            (bert-base-multilingual-cased, 68.02)
            (bert-base-multilingual-uncased, 65.98)
        };
        \end{axis}
    \end{tikzpicture}
\end{minipage}
\begin{minipage}[t]{.42\textwidth}
    \definecolor{grey}{HTML}{b4b4a9}
    \begin{tikzpicture}[scale=0.95]
        \begin{axis}[
            ybar,
            symbolic x coords={
                Co-match + W2V, 
                Co-match + W2V\_C2V, 
                Co-match + FastText, 
                Co-match + Bert\_Base, 
                Co-match + ELMO, 
                Co-match + MULTI,
                bert-base-multilingual-cased,
                bert-base-multilingual-uncased
            },
            ymin = 0, ymax = 100,
            xtick=data,
            nodes near coords,
            nodes near coords align={vertical},
            every node near coord/.append style={rotate=90, anchor=west},
    	    ylabel near ticks,
    	    x tick label style={font=\footnotesize,rotate=45, anchor=east},
        ]
        \addplot[black, fill=grey] coordinates {
            (Co-match + W2V, 37.35) 
            (Co-match + W2V\_C2V, 37.35) 
            (Co-match + FastText, 40.66) 
            (Co-match + Bert\_Base, 40.07) 
            (Co-match + ELMO, 45.17) 
            (Co-match + MULTI, 42.80)
            (bert-base-multilingual-cased, 60.50)
            (bert-base-multilingual-uncased, 61.28)
        };
        \end{axis}
    \end{tikzpicture}
\end{minipage}
\caption{\textit{Experimental results on the dev and test set, respectively}} 
\label{figure:result}
\end{figure*}

Fig. \ref{figure:result} illustrate the empirical results on the development set and test set respectively. Overall, the results from the development set and test set are just slightly different.

\subsection{Error Analysis}
In this section, we analyze the error predictions of the Co-match model and the BERT model respectively based on the reasoning types of each question. As mentioned in Section \ref{section3}, there are five reasoning types in the ViMMRC corpus: Word matching (WM), paraphrasing (P), single-sentence reasoning (SSR), multi-sentence reasoning (MSR) and ambiguous or insufficient (AB). Table \ref{table9} displays the error predictions percentage on each type of reasoning of the Co-match model and the BERT model. The percentages of error predictions are computed by dividing the total questions with the wrong answer by each reasoning types in total questions.

\begin{table}[H]
\caption{ERROR RATE BASED ON QUESTIONS REASONING TYPES OF TEST SET}
\label{table9}
\flushend
\begin{center}
\begin{threeparttable}
\begin{tabular}{|p{0.6cm}|c|c|c|c|c|c|c|}
\hline
\multirow{9}{*}{\makecell{Co-\\match}} & \textbf{\makecell{Model}} & \textbf{\makecell{WM \\ (\%)}} & \textbf{\makecell{P \\ (\%)}} & \textbf{\makecell{SSR \\ (\%)}} & \textbf{\makecell{MSR \\ (\%)}} & \textbf{\makecell{AB \\ (\%)}} & \textbf{\makecell{Total \\ (\%)}}\\
\hline
&\makecell{W2V} & 2.91 & \textbf{1.75} & 17.12 & 31.90 & 8.94 & 62.32\\
\cline{2-8}
&\makecell{W2V\_C2V} & 2.91 & \textbf{1.75} & 17.12 & 31.90 & 8.94 & 62.32\\
\cline{2-8}
&\makecell{FastText} & 2.33 & 2.52 & 17.12 & 28.79 & 8.56 & 59.32\\
\cline{2-8}
&\makecell{Bert\_Base} & 2.72 & 2.91 & 17.12 & 28.01 & 9.14 & 59.90\\
\cline{2-8}
&\makecell{ELMO} & 2.33 & 2.52 & \textbf{15.95} & \textbf{26.43} & \textbf{7.00} & \textbf{54.25} \\
\cline{2-8}
&\makecell{MULTI} & \textbf{1.94} & 2.52 & 17.12 & 27.82 & 7.78 & 57.18\\
\hline
\multirow{4}{*}{\makecell{\textcolor{black}{BERT}}} &\makecell{\textcolor{black}{\makecell{multi-\\lingual-\\uncased}}} & \textbf{1.16} & \textbf{0.58} & 10.50 & \textbf{20.23} & 6.22 & \textbf{38.69} \\
\cline{2-8}
&\makecell{\textcolor{black}{\makecell{multi-\\lingual-\\cased}}} & 1.55 & 1.36 & \textbf{9.72} & 20.81 & \textbf{6.03} & 39.47 \\
\hline
\end{tabular}
\end{threeparttable}
\begin{tablenotes}
    \small
    \item \textit{Note: \textbf{WM} - Word matching, \textbf{P} - Paraphrasing, \textbf{SSR} - Single-sentence reasoning, \textbf{MSR} - Multi-sentences reasoning, \textbf{AB} - Ambiguous or insufficient}
\end{tablenotes}
\end{center}
\end{table}

According to Table \ref{table9}, for the Co-matching model, the ELMO word embedding has the lowest error prediction rate, which is only 54.25\%. For single-sentence reasoning, multi-sentence reasoning and ambiguous or insufficient, ELMO has the lowest error rate too. However, for word matching, ELMO is not as good as MULTI, and for paraphrasing, ELMO also not as good as W2V and W2V\_C2V. However, because the proportion of single-sentence reasoning and multi-sentence reasoning in ViMMRC is larger than other types (according to Table \ref{table3} for the whole corpus and according to Table \ref{table6} for the test set), and ELMO gave the least error prediction on those two kind, according to Table \ref{table9}, thus the accuracy of ELMO is the highest result on the Co-match model. 

\textcolor{black}{On the other hand, according to Table \ref{table9}, the result of BERT model is dramatically higher than Co-match model. The bert-base-multilingual-uncased model has lower error rate than bert-base-multilingual-cased on word matching, paraphrasing and multi-sentence reasoning types. In contrast, bert-base-multilingual-cased has lower error rate on single-sentence reasoning and ambiguous or insufficient types. The total error rate of bert-base-multilingual-uncased is lower than bert-base-multilingual-cased, however, their result are not too much different. In general, the BERT model has  more optimistic result than the Co-match model on the ViMMRC corpus.}

Besides, we compare the number of words appeared on each word embedding and the accuracy of each word embedding. According to Table \ref{table7}, the appeared words (vocabulary size) of W2V and W2V\_C2V is higher than ELMO, but the accuracy is contrary. The number of appeared word in the word representation embedding does not impact to the accuracy of the model.

\section{Conclusion and Future works}
\label{section6}
In this paper, based on the multiple-choice reading comprehension for the Vietnamese corpus, we had conducted experiments with the Co-match model on six different Vietnamese word embeddings including W2V, W2V\_C2V, FastText, Bert\_Base, ELMO, and MULTI and the \textcolor{black}{two BERT model including bert-based-multilingual-cased and bert-based-multilingual-uncased. The empirical results showed three main contributions:} 
\begin{enumerate}
\item Our experiments showed that BERT proved its power because it achieved better accuracy than the Co-match model.
\item On the Co-match model, ELMO word embedding obtained the highest result. \textcolor{black}{On the BERT model, the the accuracy of the bert-based-multilingual-uncased is slightly higher than the bert-based-multilingual-cased.}
\item \textcolor{black}{Through experimental results, we obtained different results with various reasoning types of question on both the models. In particular, the large number of single-sentence and multi-sentence reasoning questions, which are difficult to find correct answers, affected to the performance of the Co-match and BERT models.}
\end{enumerate}

The machine reading comprehension task is a challenging task in natural language processing and has many applications in practice. One of the potential applications is building a chat-bot in which can understand humans' reading texts and answer the questions based on reading texts. Our works in the future include:
\begin{enumerate}
\item We will conduct experiments on the other deep neural models and transfer learning models for the multiple-choice reading comprehension in Vietnamese.
\item Based on the best-trained model and the ViMMRC corpus, we will build an intelligent chat-bot system for automatically evaluating students' reading comprehension skills, which is really helpful for primary-school students.
\item We will extend the ViMMRC corpus with more reading texts and multiple-choice questions of higher grades, which is really useful for exploring more machine-learning-based  models.
\end{enumerate}

\section*{Acknowledgment}
We would like to express our thanks to the authors and the NLP@UIT researching group for providing the very valuable corpus for our experiments. This research is funded by University of Information Technology -- Vietnam National University HoChiMinh City under grant number D1-2019-08.

\bibliographystyle{IEEEtran}
\bibliography{reference}

\end{document}